# Knowledge Retrieval for Robotic Cooking

Kundana Mandapaka
University of South Florida
kundana@usf.edu

*Abstract*— *Search algorithms are applied where data retrieval with specified specifications is required. The motivation behind developing search algorithms in Functional Object-Oriented Networks is that most of the time, a certain recipe needs to be retrieved or ingredients for a certain recipe needs to be determined. According to the introduction, there is a time when execution of an entire recipe is not available for a robot thus prompting the need to retrieve a certain recipe or ingredients. With a quality FOON, robots can decipher a task goal, find the correct objects at the required states on which to operate and output a sequence of proper manipulation motions. This paper shows several proposed weighted FOON and task planning algorithms that allow a robot and a human to successfully complete complicated tasks together with higher success rates than a human doing them alone.*

## I. INTRODUCTION

Video annotations are developed to determine the process, kitchen items, and ingredients used while making a specific recipe. These annotations are used as the basic instructions that can be used to describe the process of preparing a dish, and the state at which a kitchen item is required for preparing that specific dish is mentioned.

Every ingredient, or object present in the kitchen is described in the annotation, along with the state in which they are present, and the state in which they are required to be in so that they can be used in the cooking process. This helps the robots to easily understand the recipe and the preparation process of a particular dish.

In this project, a foon.txt file consisting of all the annotations is taken as input along with the kitchen and goal state files. Iterative Deepening Search and Greedy Best First Search algorithms are implemented to retrieve the subgraphs based on the start and the goal states.

The main aim of the project is to retrieve the subgraph of the recipe with the minimum number of inputs required and the subgraph for a recipe with the maximum success rate of a specified motion. The success rates of every motion have been specified in the motion.txt file, along with the success rate of that motion.

The previous works concentrated on developing a model that will generate a Functional Object-Oriented Networks graph. In this project we take a step forward and try to find the path from a specified start state to the mentioned goal state, giving the best possible sequence of steps.

## II. VIDEO ANNOTATIONS

Video annotation is the process used in Functional Object-Oriented Networks in describing the processes and activities happening in the video. It involves exclusive recognition and description of all objects in the video, their states, and motions/actions involved with their time frames.

Every action is referred to as motion. Each motion has its constituent inputs and outputs knowns as input nodes and output nodes respectively. The unit process which involves an action that transforms inputs through a particular motion is referred to as a functional unit.

The data retrieved from the video is organized using a tool known as a task tree. A task tree is a knowledge representation tool with a list of all objects, states, and motions. A FOON describing a sequence of processes of achieving a particular object with several functional units is referred to as a subgraph

The graph is generated using a task tree visualizer, where the nodes are connected to each other using arrows. The direction of the arrows describes the flow of action. The motion is described using red-colored squares. The objects, along with their states, are the green circles connected to the motion nodes. The goal state is described using a purple circle.

Below is an example of a functional unit subgraph describing the action of chopping an onion. The annotation describes the objects, their states, and the motion to be done to obtain the required goal, in this case, chopped onions.

| | | |
|---|---|---|
| O | onions | 1 |
| S | whole | |
| O | knife | 1 |
| S | clean | |
| O | chopping board | 0 |
| M | chop | |
| O | onions | 1 |
| S | chopped | |
| S | in [chopping board] | |
| O | knife | 1 |
| S | dirty | |
| O | chopping board | 0 |
| S | contains {chopped onion} | |

Each functional unit describes an action, the kitchen items, and the state of each item when they are being used so that it would be easy for the robot to understand the process, the items required, the state in which they are to be used, the start and the goal state.

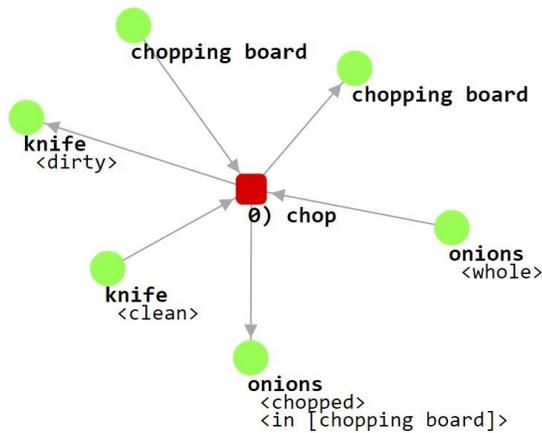

[Image showing the graph representing the task of chopping an onion]

In the above functional unit, the red-colored square represents the motion being performed while the items required are described through the green-colored circles. The goal state is described through the blue colored circle. The arrows represent the flow of motion, what items are given as input, and what is the output.

## III. METHODOLOGY

In the project the following algorithms have been implemented; Iterative Deepening Search, Greedy Best-First Search. The task of pouring ice into an empty glass is taken as a sample task.

Iterative Deepening Search is an iterative graph-searching strategy that achieves its completeness by enforcing a depth limit on DFS that mitigates the possibility of getting stuck in an infinite branch. It goes through branches of a node scanning from left to right until the required depth is achieved. Ideally, we must find the optimal solution but to make it simple, we take the first path we find. The depth is increased until the desired solution is found. A task tree is considered a solution if the leaf nodes are available in the kitchen.

The greedy best-first search algorithm is a search algorithm that makes use of the shortest path. It explores the node that is closest to the goal node. It evaluates the node using a heuristic function h(n). In the project two heuristic functions are applied to choose a path from the various options instead of choosing a path randomly.

The first heuristic functions are as follows:
- H(n) = success rate of the motion
- H(n) = number of input objects in the function unit

In the first heuristic function, the search was required to evaluate the closest node to the goal node considering the success rate value of a particular motion in the motion.txt file. In the case of this heuristic function, if we have multiple paths for a goal state with different motions, the path that gives the highest success rate of executing the motion successfully is chosen.

In the second heuristic function, the search was required to evaluate the closest node to the goal node considering the number of inputs in the functional unit. In the case of this heuristic function, if we have multiple paths to reach a goal state, the path with the least number of input objects is chosen.

| | Algorithms Used | | |
|---|---|---|---|
| Task | *Iterative Deepening Search* | *Greedy Best-First Search (success rate of motion)* | *Greedy Best-First Search (number of input objects)* |
| No.of functional units | 6 | 5 | 6 |
| Computer complexity | 60% | 80% | 60% |
| Memory complexity | HIGH | LOW | HIGH |

[Table describing the properties of the algorithms to perform a common task]

The computer complexity has been calculated based on the algorithm's success rate of the motion selected to perform a task while memory complexity has been determined based on the number of steps the path selected by the algorithm to perform a particular task contained.

## IV. DISCUSSION

From the above table, and after testing the three algorithms on multiple goal states, we can see that in the three algorithms considered, Iterative Deepening is faster than the Greedy Best- First Search algorithms but takes more memory than the Greedy best-first search algorithms. Also, upon considering the results from multiple goal states, it is observed that the performance of the Iterative Deepening Depth-First Search and the Greedy Best-First Search based on the number of input objects is almost the same.